\documentclass[conference]{IEEEtran}
\usepackage[T1]{fontenc}
\IEEEoverridecommandlockouts
\usepackage{cite}
\usepackage{amsmath,amssymb,amsfonts}
\usepackage{algorithmic}
\usepackage{graphicx}
\usepackage{textcomp}
\usepackage{xcolor}
\usepackage{hyperref}
\usepackage{float}
\def\BibTeX{{\rm B\kern-.05em{\sc i\kern-.025em b}\kern-.08em
    T\kern-.1667em\lower.7ex\hbox{E}\kern-.125emX}}
\begin{document}

\title{Pay Attention to What and Where? Interpretable Feature Extractor in Vision-based Deep Reinforcement Learning}


\author{\IEEEauthorblockN{Tien Pham {\thanks{*This work has been partially supported by Horizon Europe (and UKRI Horizon Guaranteed Fund)  under the Marie Skłodowska-Curie grant agreement No 101072488 (TRAIL). This work was also in part supported by a project funded by the EPSRC Prosperity grant CRADLE (EP/X02489X/1) and by the Air Force Office of Scientific Research, USAF, under the CASPER++ Awards (FA8655-24-1-7047).} 
\thanks{© 2025 IEEE. Personal use of this material is permitted. Permission from IEEE must be obtained for all other uses, in any current or future media, including reprinting/republishing this material for advertising or promotional purposes, creating new collective works, for resale or redistribution to servers or lists, or reuse of any copyrighted component of this work in other works.}
}}
\IEEEauthorblockA{\textit{School of Computer Science} \\
\textit{The University of Manchester}\\
Manchester, UK \\
canhantien.pham@manchester.ac.uk}
\and
\IEEEauthorblockN{Angelo Cangelosi}
\IEEEauthorblockA{\textit{School of Computer Science} \\
\textit{The University of Manchester}\\
Manchester, UK \\
angelo.cangelosi@manchester.ac.uk}
}

\maketitle

\begin{abstract}
  Current approaches in Explainable Deep Reinforcement Learning have limitations in which the attention mask has a displacement with the objects in visual input. This work addresses a spatial problem within traditional Convolutional Neural Networks (CNNs). We propose the Interpretable Feature Extractor (IFE) architecture, aimed at generating an accurate attention mask to illustrate both "what" and "where" the agent concentrates on in the spatial domain. Our design incorporates a Human-Understandable Encoding module to generate a fully interpretable attention mask, followed by an Agent-Friendly Encoding module to enhance the agent's learning efficiency. These two components together form the Interpretable Feature Extractor for vision-based deep reinforcement learning to enable the model's interpretability. The resulting attention mask is consistent, highly understandable by humans, accurate in spatial dimension, and effectively highlights important objects or locations in visual input. The Interpretable Feature Extractor is integrated into the Fast and Data-efficient Rainbow framework, and evaluated on 57 ATARI games to show the effectiveness of the proposed approach on Spatial Preservation, Interpretability, and Data-efficiency. Finally, we showcase the versatility of our approach by incorporating the IFE into the Asynchronous Advantage Actor-Critic Model.
\end{abstract}


\section{Introduction}
\label{introduction}

Deep Reinforcement Learning (DRL) has been developing rapidly and showing outstanding performance in robotics, games, and other decision-making processes. Many DRL models have outperformed human experts in fields such as AlphaGo \cite{silver2016mastering} or MEME \cite{kapturowski2022human}. However, the social impact of the DRL in reality has been underrated, especially in critical applications such as Autonomous Driving and Healthcare due to its poor interpretability. Explainable Deep Reinforcement Learning (XDRL) has been studied in many works to unveil key insights into the decision-making process of the agent. Saliency maps are generated in many works to visualize the important locations regarding the policy changes \cite{greydanus18a}. Other works \cite{mott2019towards, choi2017multi} have successfully applied a multi-head attention mechanism to reveal the task-relevant information from the visual input. All the aforementioned works can only interpret some of the games in the ATARI benchmark, especially for the games in which the model can perform better than humans but the attention maps are poorly understandable, raising the concern about the effectiveness of the approach. Furthermore, all of the proposed approaches rely on Convolutional Neural Networks to extract the environment feature from the visual input and therefore do not inherently enforce spatial consistency. We construct and evaluate a versatile \textbf{Interpretable Feature Extractor (IFE)} which can serve as a reliable extractor for vision-based deep reinforcement learning.

\begin{figure*}
  \centering
{\includegraphics[width=0.95\textwidth]{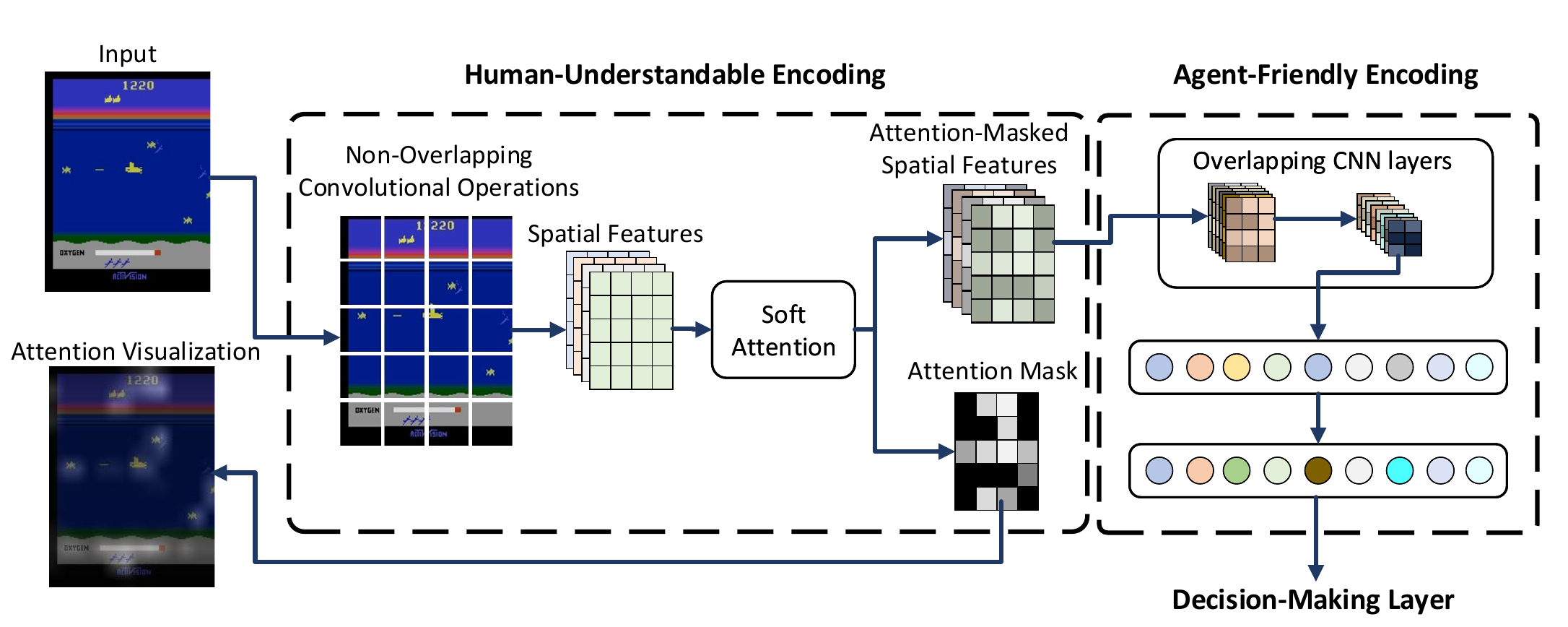}}
\vspace{-10pt}
  \caption{Proposed Interpretable Feature Extractor Architecture for Vision-based Deep Reinforcement Learning}
  \label{fig:outline_model}
\end{figure*}

In this paper, we address the spatial preservation problem in XDRL, which leads to the dilemma between interpretability and performance. On one hand, the convolutional neural network (CNN) plays an important role in DRL which efficiently extracts the spatial features from visual input. However, the spatial consistency between input and output is not fully preserved due to the overlap between convolution windows. On the other hand, non-overlapping convolutional operations can fully preserve the spatial information but will reduce the efficiency during training (Sec. \ref{subsec:spatial}). Rethinking the Feature Extractor in DRL, an interpretable model is constructed that balances spatial preservation and learning performance (illustrated in Fig. \ref{fig:outline_model}). Firstly, this approach uses non-overlapping convolutional operation to extract the features from visual input while fully preserving the spatial information, followed by a soft attention mechanism to force the model to accurately focus only on the important features relating to the decision during training. The attention mask is computed on the forward pass of the model during inference and is in a single map, which is more efficient for visualization, and consistent with the context. After that, the features will be transformed into an agent-friendly domain, which allows the model to flexibly encode the environment representation into the feature maps that are efficient for the model to learn. As a result, the feature extractor is interpretable and efficient for the Reinforcement Learning agent to learn regardless of the learning type or policy. We show that our Interpretable Feature Extractor could produce an accurate, consistent, and highly human-understandable attention mask from all 57 games in ATARI environment \footnote{Videos of all ATARI environments can be found at: \href{https://sites.google.com/view/pay-attention-to-windows}{https://sites.google.com/view/pay-attention-to-windows} \label{link:webpage}}. To promote the reproducibility and reusability, our code is publicly available \footnote{\href{https://github.com/tiencapham/IFE}{https://github.com/tiencapham/IFE}}.

We summarize our contributions as follows:
\begin{itemize}
    \item Address the spatial problem in Explainable AI and propose an easy but effective approach to minimize that problem in Vision-based Deep Reinforcement Learning.
    \item Introduce a versatile and reliable feature extractor that is integrated into Deep Reinforcement Learning Models to enable interpretability. 
    \item Evaluate the approach on Spatial Preservation, Interpretability, Data-efficient against the traditional CNN approach and the SOTA Interpretable Deep Reinforcement Learning model (Mott et al. \cite{mott2019towards}) on ATARI environment.
\end{itemize}

\section{Preliminaries}
\label{sec:prelim}

\subsection{Model-free Deep Reinforcement Learning}

We consider Atari Gameplay to be a Markov Decision Problem defined by tuple $(\mathcal{S}, \mathcal{A}, \mathcal{P}, \mathcal{R}, \gamma)$. The State Space $\mathcal{S}$ includes all possible states $s$ presented by visual inputs from the game engine. $\mathcal{A}$ is a finite set of possible discrete actions $a$ that the agent could perform given each state. $\mathcal{P}$ is the state-transition distribution which indicates the probability of state-transition for each tuple $(s, a)$. The reward function $r$ is an output provided by the environment while performing a specific action in a state, and the discount factor $\gamma$ reflects the weight of the reward signal over time. The agent, by performing a series of actions based on the visual states to collect the rewards, will try to explore the environment and maximize the reward gained via training. There are several approaches to tackling the large and possibly infinite state spaces that try to estimate the expected sum of future reward when taking the action following the optimal policy thereafter \cite{mnih2016asynchronous, van2016deep, mnih2015human, hessel2018rainbow, assran2019gossip}. Several frameworks have been developed to improve the performance and learning efficiency of the agent such as Deep Q-Network \cite{van2016deep}, Proximal Policy Optimization \cite{schulman2017proximal}, Actor Critics \cite{lee2020stochastic}, Asynchronous Advance Actor-Critic \cite{mnih2016asynchronous}, Impala \cite{espeholt2018impala}, MEME \cite{kapturowski2022human}, etc. All of those approaches use the Convolutional Neural Network Variant to extract the features that represent the visual state which is then mapped with the actions to indicate the state-action pair vector.
Our work focuses on improving the interpretability of the feature extractor by explaining the perception of the agent regarding the environment which could be applied in most of the vision-based deep reinforcement learning models. However, due to the time-consuming training in deep reinforcement learning, we only integrate the approach on Fast and Data-efficient Rainbow \cite{schmidt2021fast} which serves as the main framework for our evaluation due to its data efficiency in learning.
We also extend the approach to Asynchronous Advantage Actor-Critic (A3C) with Long-Short Term Memory (LSTM) \cite{mnih2016asynchronous}, which has a different network structure as well as learning policy and observation configuration, to prove its versatility.

\subsection{Attention}
\label{subsec:attention}

Attention mechanism has been well studied recently in multiple domains \cite{bahdanau2014neural,xu2015show, caron2021emerging, mott2019towards}. The attention 
to visual input will indicate the locations on the image on which the model is focusing to decide the action. The key idea is to apply a learnable attention weight over the spatial features, allowing the model to adapt to focus on relevant spatial features in the decision-making process. There are two main types of attention mechanisms namely Multi-head Attention \cite{vaswani2017attention} and Bahdanau Attention \cite{bahdanau2014neural}.
\textbf{Multi-head attention} simultaneously creates multiple mappings between a query $\mathcal{Q}$ and a set of key-value $(\mathcal{K}, \mathcal{V})$ pairs to an input, producing the low dimensional vectors summarizing the input based on the query \cite{vaswani2017attention}. Each mapping will create the attention map, together representing the attention of the model in different spaces.
By contrast, \textbf{Bahdanau Attention} transforms the input into attention space over the spatial dimensions, then bottlenecked into a single value to create the attention weights. There are also other forms and variations of the attention mechanisms that has been used intensively to improve the model performance as well as increase the interpretability of the model \cite{woo2018cbam, yan2019stat}. This work uses a soft mechanism in Bahdanau Attention (Sec. \ref{sec:method}) to produce the attention mask due to its training efficiency as well as the interpretability of the mask. 

\subsection{Spatial Preservation versus Performance in CNN}
\label{subsec:spatial}

\begin{figure}[H]
\centerline{\includegraphics[width=0.4\textwidth]{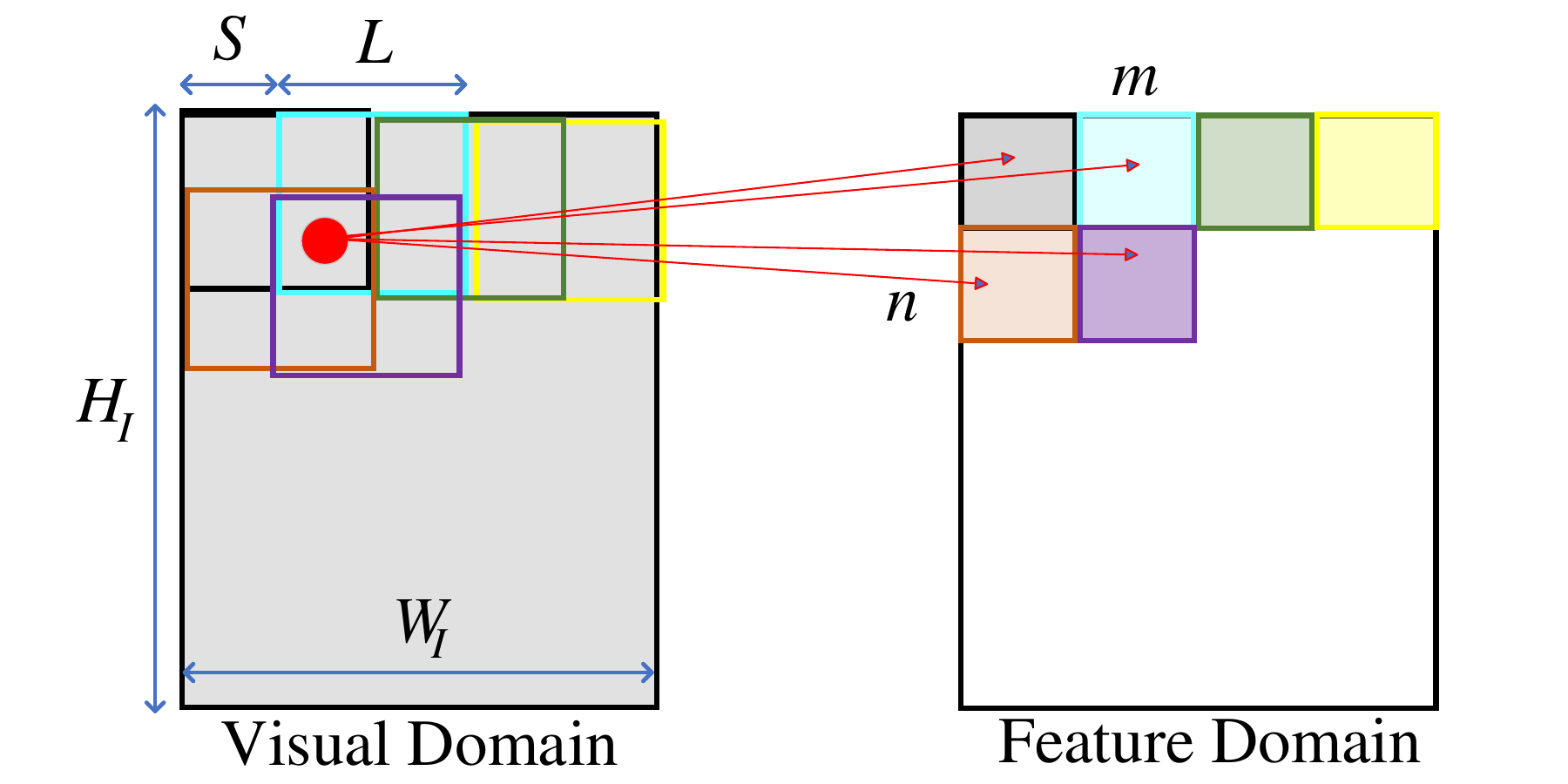}}
  \caption{Illustration of overlapping convolutional operations result in one-to-many transformation.}
  \label{fig:one-to-many}
\end{figure}

Traditional Vision-based Deep Reinforcement Learning relies on Convolutional Neural Networks as a feature extractor to gradually extract the visual features in \textbf{Overlapping Spatial Domain} \cite{Goodfellow-et-al-2016}. In many works on XDRL \cite{mott2019towards,greydanus18a, choi2017multi, sorokin2015deep}, the visual explanation such as attention or saliency map is computed in the final features and transformed into visual space. However, convolutional operation is non-invertible which does not allow an accurate transformation from feature to visual domain. As a common practice, a simple upsampling approach is used to retrieve the attention mask in visual input. Assuming a feature has coordinates $(m,n)$ in the feature domain and is extracted by the Convolution Kernel size $(L, L)$ with the stride $S$ from the visual input that has dimensions of $(W_{I},H_{I})$. After the upsampling process, the displacement between the attention mask from this feature and its corresponding pixels in Visual Input is defined as follows:

 \begin{equation}
    \begin{cases}
D_{x} = mS(1-\frac{1}{1+\frac{S-L}{W_{I}}}) +l_{x}(1-\frac{S}{L}\frac{1}{1+\frac{S-L}{W_{I}}}) \\
D_{y} = nS(1-\frac{1}{1+\frac{S-L}{H_{I}}}) +l_{y}(1-\frac{S}{L}\frac{1}{1+\frac{S-L}{H_{I}}})
\label{equa:displacement_main}
\end{cases}
\end{equation}

As a result, a \textbf{spatial problem}, where there are displacements between the attention mask and its corresponding visual input, is unavoidable in Explainable Vision-based Deep Reinforcement Learning using the traditional CNN approach. Additionally, the \textbf{one-to-many problem}, where the object located in overlapping regions could be extracted by multiple convolution windows, is also caused by the overlapping convolutional operations. As a result, there are possibly multiple features of the same objects (Fig. \ref{fig:one-to-many}).
In contrast, a non-overlapping convolution, where $S$ is equal to $L$, is fully spatial preservable which could produce an accurate attention mask. However, this approach loses 
the \textbf{parameter sharing} \cite{Goodfellow-et-al-2016} between convolutional operations, resulting in a reduction of learning performance. The parameter sharing allows the model to define which domain the features are extracted, giving the transformation flexibility to learn and adapt to the task respectively. Therefore, a dilemma between interpretability and performance needs to be balanced to achieve Explainable Deep Reinforcement Learning.

\section{Related Works}

\textbf{Attention mechanism} has been well-studied in Deep Learning to improve the interpretability and performance of the models, such as text translation \cite{bahdanau2014neural, vaswani2017attention}, image captioning \cite{xu2015show, li2017mam,hudson2018compositional}, question answering \cite{hermann2015teaching, mnih2014recurrent}, object tracking \cite{kosiorek2017hierarchical} and reinforcement learning \cite{choi2017multi, sorokin2015deep, manchin2019reinforcement, mott2019towards}. Those approaches use attention masks to provide the perception of the models regarding the input that highlights the most task-relevant information. For vision-related tasks, all of the aforementioned works use a Convolutional Neural Network as a backbone to extract the visual features and apply the attention mechanism to create the masks highlighting the locations with high attention weight. However, we observe that the CNNs cause the spatial problem mentioned in Sec. \ref{subsec:spatial}. To the best of our knowledge, we are the first to address and solve the spatial problem in the vision-based attention model. 

\textbf{Interpretable Deep Reinforcement Learning.} Interpretability of the RL model could be explored by several aspects, from self-interpretable modeling to reward decomposition and post-training explanation \cite{guo2021edge}. The reward decomposition \cite{lin2020contrastive, juozapaitis2019explainable} could be used to interpret the action of the agent based on the re-engineered reward. The saliency map \cite{atrey2019exploratory, puri2019explain} could be extracted from the traditional RL model to explain the agent's decision after training. Other works \cite{guo2021edge, puri2019explain} identify the critical states for the final rewards of the agent. Our work focuses on enabling the interpretability of the agent by designing an interpretable feature extractor.

\textbf{Attention in RL.} There have been several works that applied attention mechanisms to unveil the information in Deep Reinforcement Learning during the decision-making process \cite{sorokin2015deep, choi2017multi, mott2019towards}. Sorokin et al. \cite{sorokin2015deep} proposed a Deep Attention Recurrent Q-Network combining soft or hard attention mechanisms with LSTM to extract the time-dependant attention-weighted features. However, the attention visualization is vague and does not accurately focus on the important objects. Shi et al. \cite{9259236} proposed an interpretable feature extractor by self-supervised training of the network with a decoder, producing an attention map to indicate the agent focus. Shi et al. \cite{9150853} design an attention bottleneck model to concatenate the latent representation of the network. Choi et al. \cite{choi2017multi} incorporate multi-head attention with LSTM to create a Multi-focus Attention Network (MANet) to improve the agent's capability to focus on significant elements through the utilization of multiple concurrent attention mechanisms. A similar approach was also proposed by Mott et al. \cite{mott2019towards} which apply a Soft, Spatial Sequential, Top-Down Attention (S3TA) to generate the attention masks in ATARI environment and unveil some important information underlying the decision-making. The two works combine a multi-head attention mechanism with LSTM layer to create the query vector and extract the attention from small informative regions. The multi-head attention creates multiple attention heads which represent different perceptions of the model towards the input, thus creating the inconsistency in interpretability between heads. In contrast, our soft attention creates only one attention mask which achieves better interpretability evaluated in Sec. \ref{subsec:eval_inter}.
Additionally, our approach has a minimal modification on the network and training procedure, and does not rely on the LSTM layer which could be integrated into multiple types of Reinforcement Learning models without adding new components.

\section{Method}
\label{sec:method}
Recognizing the dilemma mentioned in Sec. \ref{subsec:spatial}, we formulate the interpretable feature extractor as follows:
\begin{itemize}
    \item \textbf{Interpretability in Human Understandable Domain:} The purpose of Explainable Deep Reinforcement Learning is to generate a Visual "Explanation" understood by humans during the decision-making process. Therefore, the attention mask needs to be produced in a Human-Understandable Domain. At this stage, the model encodes the visual input into a feature domain where the semantics should be spatially maintained.
    \item \textbf{Performance in Agent-Friendly Domain:} The output of the feature extractor needs to be used by the agent to predict the action that optimizes the future reward. To improve the data efficiency of a reinforcement learning model, the output feature should be in an Agent-Friendly Domain. This allows the model to be flexible in converting the features into the domain where it promotes training efficiency.
\end{itemize}


\subsection{Human-Understandable Encoding (HUE)}

We introduce the first module in Interpretable Feature Extractor named \textbf{Human-Understandable Encoding} which primarily focuses on generating the visual explanation. An observation (a stack of grayscale frames) is processed through non-overlapping convolutional layers to extract the important features in the fully spatial preservable domain. The output tensor $[z_{i}]$ has a lower resolution of the visual input, which reduces the computational intensity of the attention process but is still fine-grained enough to produce an accurate and sharp attention mask. First, the feature tensor $[z_{i}]$ is permuted to flatten the two spatial dimensions, accelerating the transformation.
We define a learnable transformation $\phi$ that converts the vectors representing the features extracted in spatial locations into attention weights $[\alpha_{i}]$ which indicates the importance of features relating to the tasks. The feature vectors are then masked by the attention which only exaggerates the feature in important locations while reducing the values of non-related spatial features, producing attention-weighted spatial features $z_{i}^{masked}$.
\begin{equation}
\alpha_{i} = \phi(z_{i})
\label{equa:attention}
\end{equation}
\begin{equation}
z_{i}^{masked} = z_{i}\alpha_{i}
\label{equa:attention_mask}
\end{equation}
The attention transformation $f_{att}$ is constructed by two fully connected layers. The first layer is used to transform the feature vector ${z_{i}}$ into an attention domain $A$. Subsequently, the second layer converts this to a single value representing the importance weights $e_{i}$. To force the model only selecting the truly important features, a soft-max activation is applied, which results in a clear and interpretable attention mask.

\begin{equation}
e_{i} = f_{att}(z_i)
\label{equa:attention_1}
\end{equation}
\begin{equation}
\phi(z_{i})=\frac{\exp(e_{i})}{\sum_{k=1}^{L}(\exp(e_{k}))}
\label{equa:attention_2}
\end{equation}

This mechanism provides a learnable method to adjust the features by increasing the weight of important spatial features and reducing the weight of non-relating ones. Finally, the Attention-Weighted Spatial Features are permuted back to the original shape for the next processing. It is worth noticing that Human-Understandable Encoding is fully differentiable due to the combination of the convolutional layers and attention mechanism which could be trained via back-propagation. The Reinforcement Learning Model, while updating its weights to maximize future return, will also adjust the attention mechanism to select the important spatial features that contribute to the decision-making process. As a result, the Attention-Weighted Spatial Features only contain high values in important locations and negligible weights otherwise. Besides, the attention mask, where the spatial characteristic is maintained, could be overlayed on the visual input and provide the visual explanation of "what" and "where" the model is focusing on.
\subsection{Agent-Friendly Encoding}
While Human-understandable Encoding could extract the feature from visual input and create an accurate attention mask highlighting the important locations, its lack of parameter sharing could reduce the training efficiency which is the advantage of the Convolutional Feature Extractor. We introduce the second component, namely \textbf{Agent Friendly Encoding}, allowing parameter sharing in feature extraction and facilitating the training process. Convolutional layers are used in an overlapping manner followed by the activation functions, pooling or normalization layers, and other neural network variations. There is no constraint or specific requirement on the network type or architecture to be used in this module, as long as it facilitates the training efficiency of the overall network. Therefore, this approach could be adopted by multiple types of Deep Reinforcement Learning Model. Our work uses the IMPALA-Large model \cite{espeholt2018impala} as the main framework which uses a Max Pooling layer, two consecutive Residual Blocks, and an Adaptive Max Pooling Layer in the Agent-Friendly Encoding. We also integrate the approach onto A3C-LSTM which uses two convolutional layers with a fully connected layer followed by an LSTM layer.

\subsection{Interpretable Feature Extractor}
The two aforementioned components together form an \textbf{Interpretable Feature Extractor (IFE)} which could be used in a Vision-based Deep Reinforcement Learning Model. By combining the IFE with the \textbf{Decision-Making Layer} depending on the type of Reinforcement Learning Model such as Dueling or Actor-Critic Network, the model becomes interpretable. The interpretable model is able to produce the attention mask on the spatial domain, and agent-friendly features to be effectively involved in the decision-making process.

\section{Results and Analyses}
\label{sec:eval}

\begin{figure}[t]
    \centering
    {\includegraphics[width=0.48\textwidth]{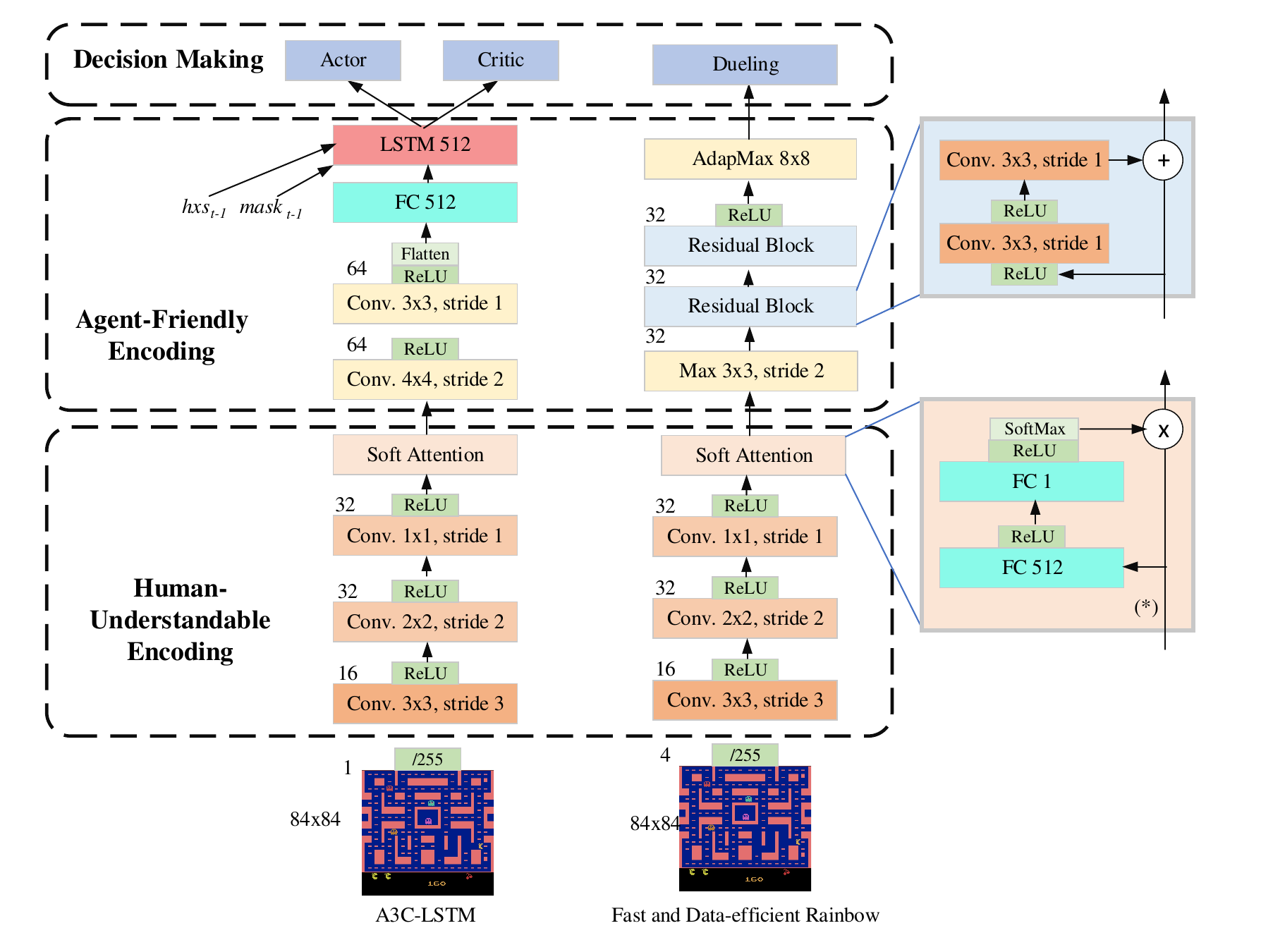}} 
    \vspace{-5pt}
    \caption{Network Architectures used in the paper}
    \label{fig:network}
\end{figure}

\begin{table}[h]
  \caption{Hyperparameters for training}
  \label{tab:rainbow_hyper}
  \centering
  \begin{tabular}{lcc}
    \hline
    \textbf{Parameter}     & \textbf{Rainbow}  & \textbf{A3C-LSTM}   \\
    \hline
    Discount factor $\gamma$ & 0.99  & 0.99  \\
    Q-target update frequency     & 32,000 frames  & -- \\
    Importance sampling $\beta_{0}$ for PER     &  0.45   &  --    \\
    $n$ in n-step bootstrapping     & 3    &  20   \\
    Initial exploration $\epsilon$    &  1    &  --   \\
    Final  exploration $\epsilon$    &  0.01    &  --   \\
    Exploration decay time & 1,000,000 frames  & --\\
    Learning rate & 0.00025  & 0.0001 \\
    Optimizer & Adam  & Adam \\
    Adam parameter & 0.005/batch size & Use Amsgrad \\
    GAE Parameter & -- & 0.92 \\
    Gradient clip norm & 10  &  -- \\
    Loss function & Huber  & --\\
    Batchsize & 256 &  --\\
    \hline
    Frameskip & 4  &  4\\
    Framestack & 4 &  1\\
    Grayscale & Yes  & Yes\\
    \hline
  \end{tabular}
\end{table}
A thorough evaluation is performed on 57 ATARI Games (OpenAI Gym v0.18.0 \cite{1606.01540}) based on our proposed approach integrated into the Fast and Efficient Rainbow model \cite{schmidt2021fast}. Details about the network architectures used in the evaluation are shown in Fig. \ref{fig:network}. Due to the space limitation of the paper, we only show a few examples for each evaluation, for a thorough understanding of how the interpretable attention masks are visualized, we would suggest visiting our project webpage\textsuperscript{\ref{link:webpage}}. Additionally, we use the Multi-head Attention Model (S3TA) from \cite{mott2019towards} as the baseline for our evaluation as they are considered as the SOTA approach for Interpretable Deep Reinforcement Learning with similar settings \footnote{The visualization of S3TA can be found at \href{https://sites.google.com/view/s3ta}{https://sites.google.com/view/s3ta}.}.

\begin{figure}[t]
    \centering
    {\includegraphics[width=0.48\textwidth]{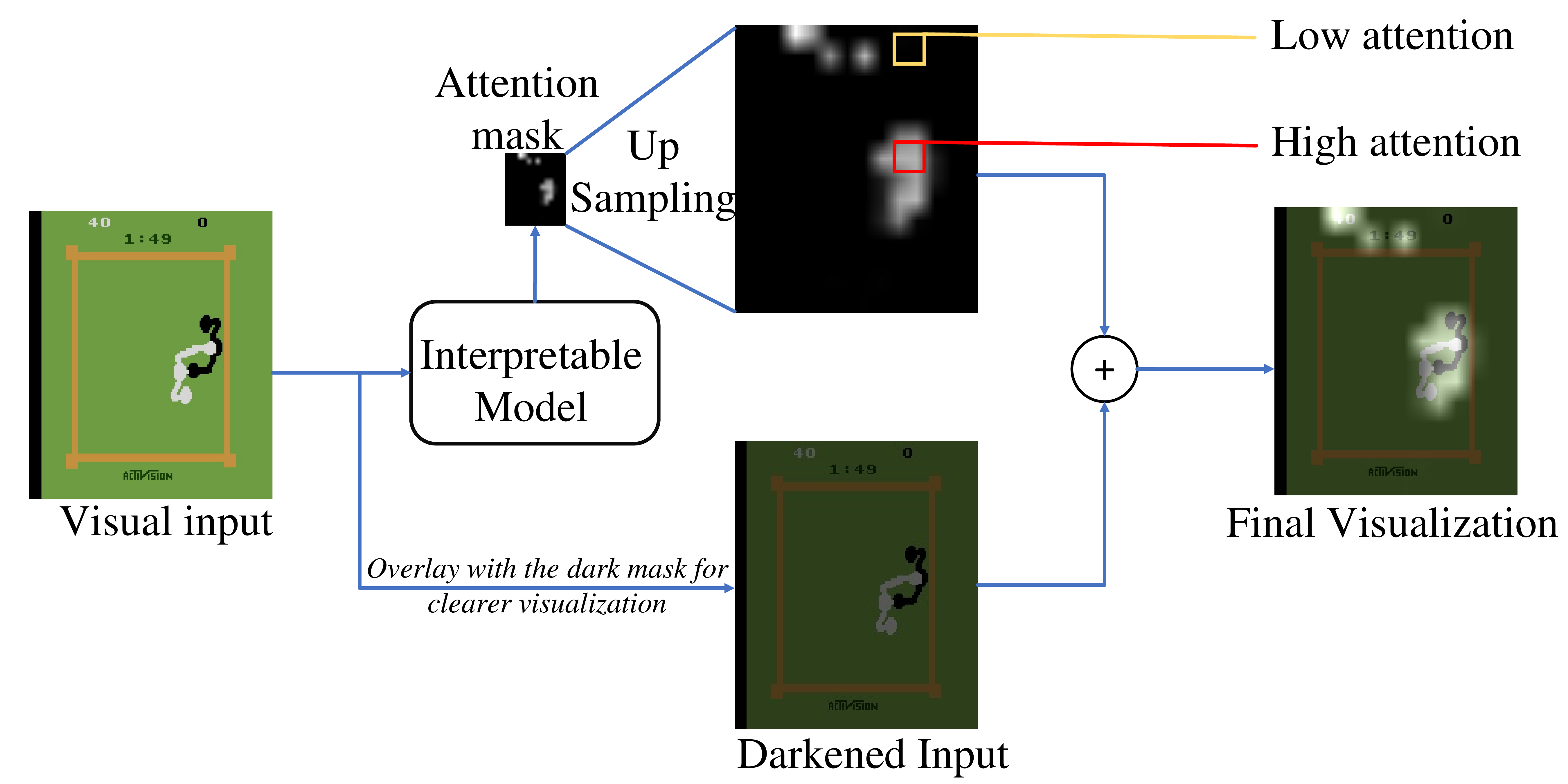}} 
    \vspace{-5pt}
    \caption{Visualization of the attention mask overlayed on visual input}
    \label{fig:visualization}
\end{figure}

\textbf{Attention Visualization.} To visualize the attention mask during the inference, we first input the visual observation into the interpretable model and retrieve the attention mask (illustrated in Fig. \ref{fig:visualization}). The mask is then upsampled to have a similar shape to the visual input. Finally, the attention mask is overlaid on the darkened input for clearer visualization. Dark attention shows a small value while the bright mask illustrates substantial attention weights highlighting the "where" and "what" the agent is looking at.  

\subsection{Spatial Preservation}
\label{subsec:spatial_preserv}

\begin{figure}[h]
    {\includegraphics[width=0.48\textwidth]{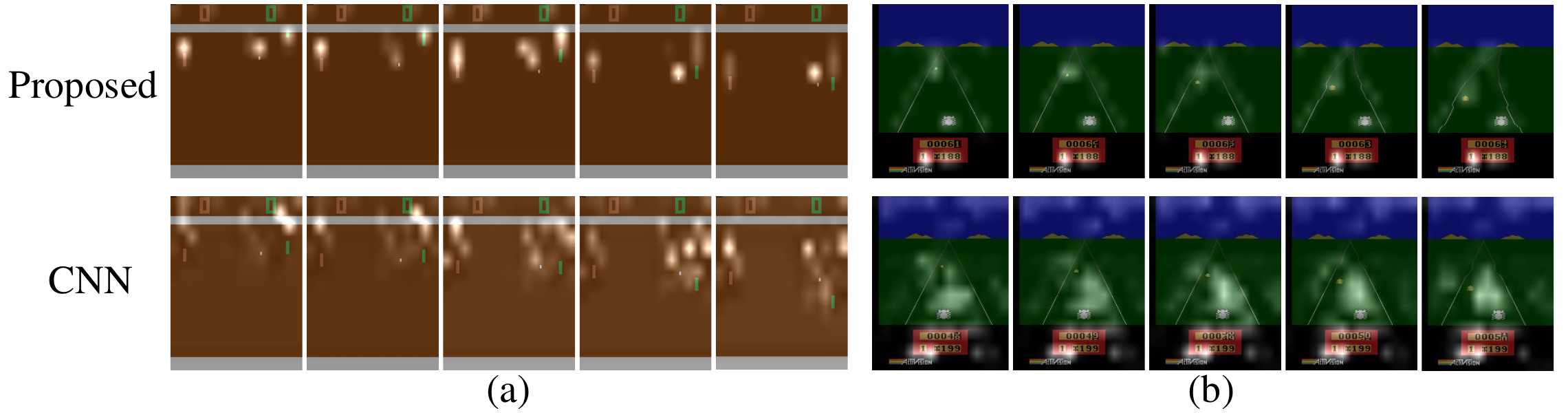}} 
    \vspace{-10pt}
    \caption{Example of 5 consecutive inputs of the model with the attention mask overlay. The figures show the comparison between the proposed model with the traditional CNN model in (a) Pong and (b) Enduro. The attention visualization of the proposed model is much clearer and more interpretable, while that of the CNN approach is distorted and blurred by multiple attention masks for the object. We can also see the attention consistent with the movement of the objects such as the ball or the car.}
    \label{fig:spatial_preservation}
\end{figure}

We compare the attention maps produced by our proposed model against the normal CNN architecture (using overlapping convolution operations in HUE). Our model can accurately localize the important objects while there are random shifts in CNN attention (See Fig. \ref{fig:spatial_preservation}). Note that the input of the state $S_{t}$ is the 4 consecutive frames $(F_{t-3}, F_{t-2}, F_{t-1}, F_{t})$ which causes the displacement of the objects in visual input, leading to the trajectory attention in the mask. Out of 57 ATARI games, we observe that our proposed model can produce clear and accurate attention masks in all environments regardless of the scoring performance. This observation supports the claim that the proposed framework mitigates the spatial preservation problem compared to the traditional CNN approach.

\begin{figure}[h]
    \centering
{\includegraphics[width=0.48\textwidth]
{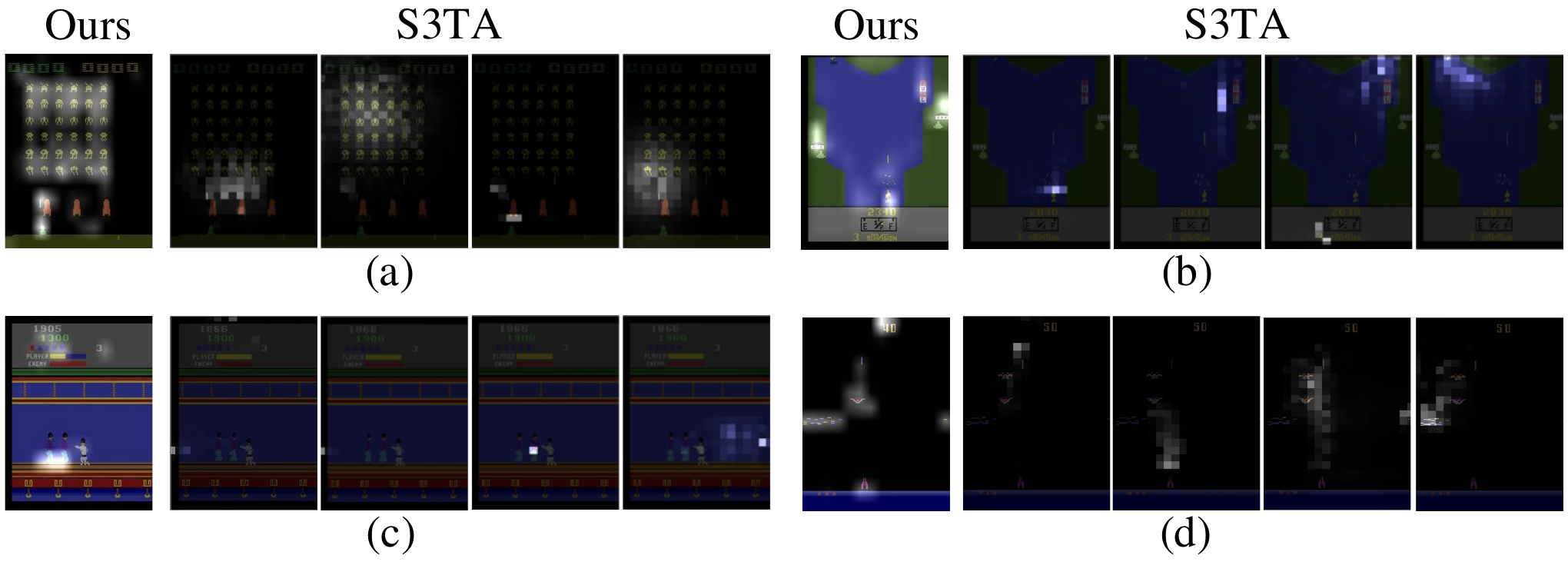}} 
\vspace{-5pt}
\caption{Attention Masks Comparision between our work and S3TA \cite{mott2019towards} in (a) Space Invaders, (b) Riverraid, (c) Kungfu Master, and (d) Assault. Four attention heads of S3TA are presented in left-to-right order.}
    \label{fig:s3ta_eval}
\end{figure}

Comparing the S3TA approach shown in Fig. \ref{fig:s3ta_eval}, we found that our attention mask is accurate and consistent with the environment context while there is a shift from S3TA visualization due to the spatial preservation problem. Taking Space Invaders (Fig. \ref{fig:s3ta_eval}(a)) as an example, the attention mask on enemies is shifted to the top-left in attention head 2, but to the bottom-left in attention head 4. In the Riverraid case (Fig. \ref{fig:s3ta_eval}(b)), the attention on the Fuel Tank is also shifted to the left in attention head 2, while multiple attentions on that tank are found in attention head 3 with random shifts from left to top. A similar problem could also be found in S3TA visualization on other games but not in our proposed approach. Furthermore, the inconsistency of attention heads also appears in the S3TA model. For example, the attention on the character could appear on attention head 3 in Space Invaders (Fig. \ref{fig:s3ta_eval} (a)) or attention head 1 in Riverraid (Fig. \ref{fig:s3ta_eval} (b)), which could raise the concern of the model interpretation. Our approach produces all the attention in a single mask, leading to more computationally efficient and approachable interpretations.

\subsection{Interpretability Evaluation}
\label{subsec:eval_inter}
Interpretability is difficult to measure and depends on various subjective and objective factors. There is currently no standard benchmark to evaluate the interpretability of the model. However,
we could intuitively claim that our interpretable model produces clear and accurate attention masks for all of the games in ATARI benchmark. 
\begin{figure}[h]
    \centering
    {\includegraphics[width=0.48\textwidth]{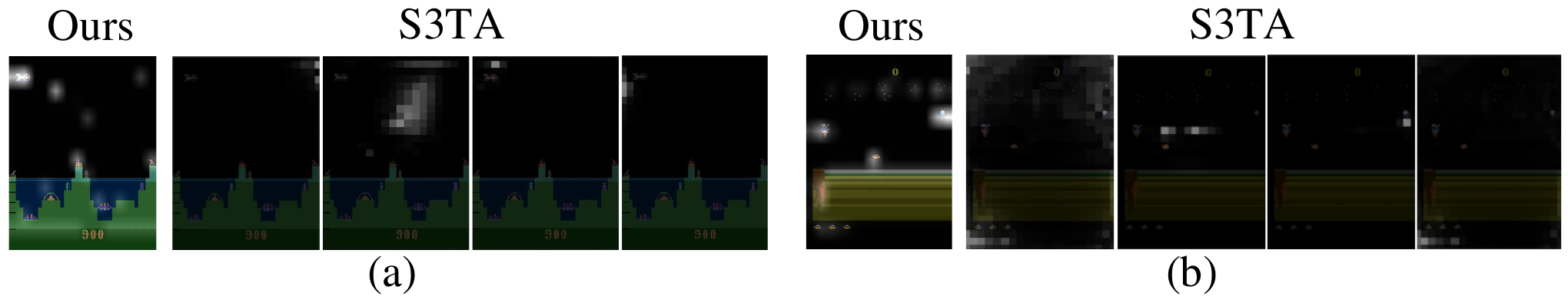}} 
    \vspace{-5pt}
    \caption{Attention Masks Comparision between our work and S3TA \cite{mott2019towards} in (a) Atlantis, (b) Jamesbond. There are random and uninterpretable attention masks captured in the S3TA approach (attention 2 in Atlantis and attention 1 in Jamesbond)}
    \label{fig:inter_eval}
\end{figure}

Examining the S3TA visualization, we observe that the displacement between attention and objects, as well as the hard-to-understand attention, appears in most of the games (Fig. \ref{fig:s3ta_eval} and Fig. \ref{fig:inter_eval}), reducing the interpretability of the visualization. Similar problems could also be seen in the CNN network with attention (Fig. \ref{fig:spatial_preservation}). This could be explained by the spatial and one-to-many problems in CNN mentioned in Sec. \ref{subsec:spatial} that blurs the attention masks and generates uninterpretable attention in the visual domain. By contrast, the proposed approach has mitigated that problem by forcing the model to encode the image in a fully-preserved spatial domain and produce a sharp, accurate, and interpretable attention mask, confirming the superior interpretability of the proposed approach.

\begin{figure}[h]
    \centering
    {\includegraphics[width=0.48\textwidth]{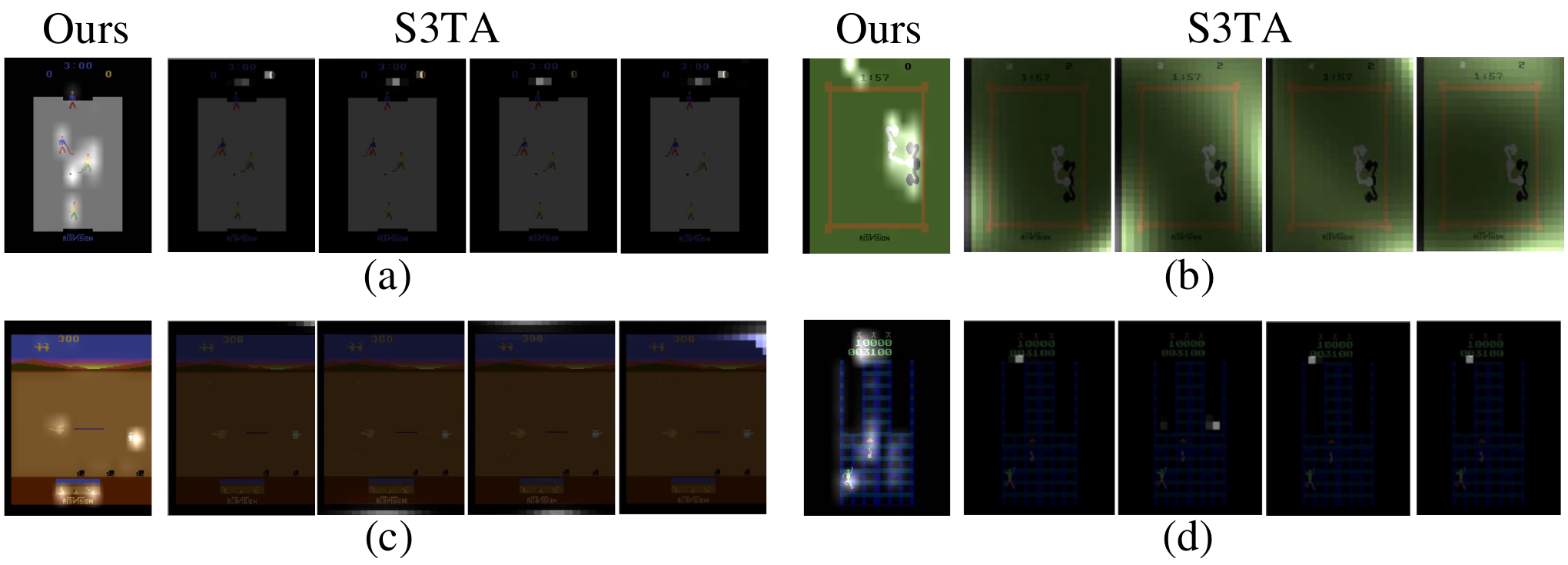}} 
    \vspace{-5pt}
    \caption{Attention Masks Comparison between our work and S3TA \cite{mott2019towards} in (a) Ice Hockey, (b) Boxing, (c) Chopper Command, and (d) Crazy Climber. We show four examples of games in which the attention mask of the S3TA model is hard to understand while Human Normalized Score $>10\%$}
    \label{fig:uninter_eval}
\end{figure}

\textbf{Interpretability in successful learning environment}. By observing all the visualizations in ATARI benchmark, we conclude that our model is interpretable for all the environments but not the case in the S3TA model. We observe environments that the S3TA model can learn with the Human Normalized Score $>10\%$ but its visualization is hard to understand (the attention mask is generated randomly, difficult to link between the attention mask and the visual input context, see Fig. \ref{fig:uninter_eval}). This phenomenon is caused by the severe spatial problem in CNN where the agent-friendly domain is extremely different from the human-understandable domain. As a result, we cannot interpret the attention mask which should have the meaning for the agent to gain the reward.

\subsection{Data efficiency in training and versatility}

We evaluate the data efficiency of the proposed model against the traditional CNN with an attention approach and one model variation, where only HUE is used for the feature extractor. The models are trained 
for $50,000,000$ frames with the benchmark performance shown in Table \ref{tab:hns_score}. We found that our proposed model has comparable performance to the traditional CNN, while the HUE-only model has the worse performance due to the lack of parameter sharing mentioned in Sec. \ref{subsec:spatial_preserv}. Details of performances and learning curves are presented in our project web page \textsuperscript{\ref{link:webpage}}. Due to the lack of open-source code of 
S3TA as well as its high computing requirement for training ($200,000,000$ frames), we compared the efficiency of our proposed approach on only 6 ATARI games including Pong, Space Invaders, Boxing, Ms Pacman, Defender, and Breakout. Our proposed approach outperformed S3TA in both mean and median metrics.

\begin{table}[h]
  \caption{Human Normalized Scores on Atari}
  \label{tab:hns_score}
  \centering
  \begin{tabular}{lll}
   \hline
    Model     & Median     & Mean \\
    \hline
    Baseline [57 games] &  922.43\%   &  139.75\% \\
    CNN with attention [57 games] & \textbf{955.48\%}  & $145.99\%$    \\
    HUIE-only [57 games]    & $896.16\%$ & $133.52\%$     \\
    Proposed [57 games]    &  $944.36\%$      & \textbf{157.21\%} \\
    \hline
    S3TA [6 games]  & 1513.4 \% & 1796.6 \% \\
    Proposed [6 games] & \textbf{1549.71 \%} & \textbf{1969.82} \%\\
    \hline
  \end{tabular}
\end{table}

\begin{figure}[h]
    \centering
    {\includegraphics[width=0.48\textwidth]{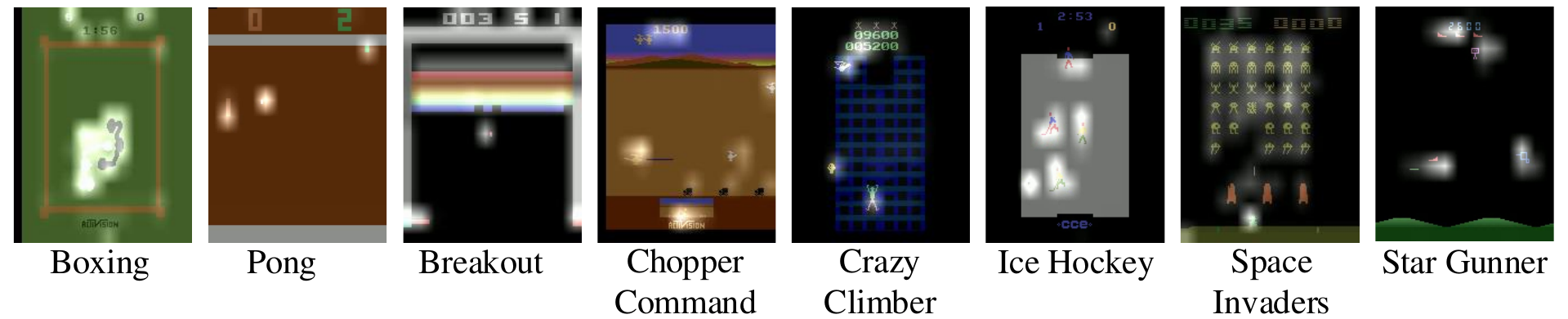}} 
    \vspace{-4pt}
    \caption{Showcases of Interpretable Feature Extractor integrated to A3C LSTM}
    \label{fig:a3c_eval}
\end{figure}

To evaluate the versatility of Interpretable Feature Extractor, we integrate the IFE into the A3C LSTM framework which has different learning categories (on-policy), different observation configurations (single grayscale frame), and different network components (LSTM Layer) compared to Fast and Efficient-Rainbow \cite{sutton2018reinforcement}. We observe a similar visualization where the attention mask accurately highlights the important object in visual input, which successfully shows "where" and "what" the agent perceives to decide the action (shown in Fig. \ref{fig:a3c_eval}).

\subsection{Attention in Transfer Learning}
\begin{figure}[h]
    \centering
    {\includegraphics[width=0.48\textwidth]{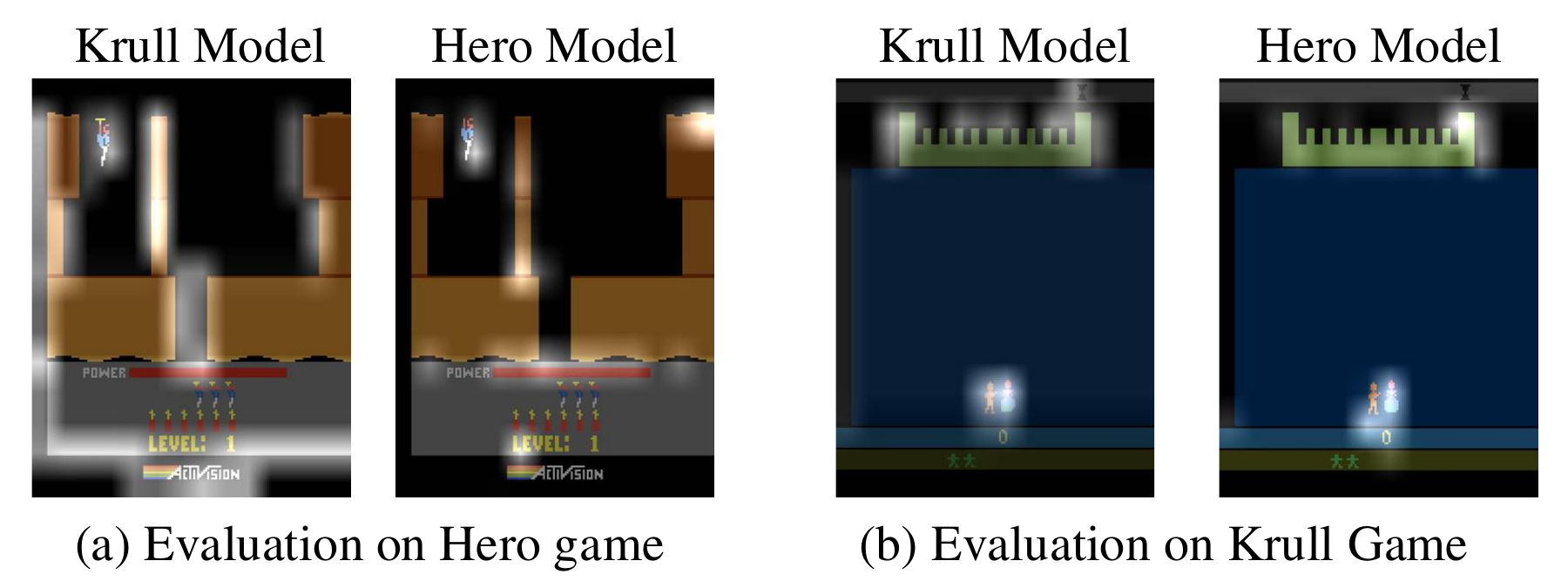}} 
    \vspace{-8pt}
    \caption{Transfer Knowledge of Interpretable Feature Extractor}
    \label{fig:transfer}
\end{figure}

Transfer Learning or continuous learning is the type of learning where we leverage the training knowledge in one task to transfer to another task, which has been emerging recently in various studies \cite{pmlr-v199-powers22b, NEURIPS2023_9d8cf124, NEURIPS2022_2938ad04}. We conducted an experiment on how attention map is transferred in the context of continual learning on Krull and Hero games, which are used to evaluate Continual Learning in ATARI benchmark \cite{NEURIPS2019_fa7cdfad}. We used the model trained in one game to evaluate in another game and found that the attention mask could also be transferred (Fig. \ref{fig:transfer}), which indicates the interpretable feature extractor partially transferred the knowledge from an encoder. This result also illustrates that the attention mask is still reliable given the unknown observation during training.

\section{Limitations}
\label{sec:limt}
While our approach has shown promising results and proven to be superior to the current state-of-the-art in Interpretable Deep Reinforcement Learning, we acknowledge several limitations in our work. Due to resource constraints and the computational intensity of Deep Reinforcement Learning, we were able to evaluate our approach only on the Fast and Data-efficient Rainbow framework and a few environments using A3C LSTM. Consequently, aspects such as data-efficient learning in Interpretable A3C LSTM, a thorough comparison of our approach's data efficiency with S3TA, and its performance in other Reinforcement Learning frameworks were not included in our evaluation. Additionally, we noticed that in some ATARI environments, the model could become trapped in sub-optimal solutions (e.g., doing nothing, resulting in zero reward). This causes the environment to remain static, providing no new visual input and therefore no useful information for learning, which leads to blurred attention visualizations. To mitigate this issue, we had to run multiple experiments on games like Tennis and Montezuma's Revenge.

\section{Conclusion}
\label{sec:conclusion}
We have successfully addressed, formulated, and developed an approach to mitigate the spatial preservation problem in vision-based Interpretable Deep Reinforcement Learning. Our approach incorporates Human-Understandable Encoding combined with a Soft Attention module to extract spatial attention from visual input, followed by Agent-Friendly Encoding to enhance the model's training efficiency. Together, these elements form an interpretable feature extractor.

Our proposed method can produce an attention mask that visualizes the agent's perception of the visual input during the decision-making process. We have evaluated and demonstrated the superiority of this approach in terms of spatial preservation, interpretability, and data efficiency compared to traditional CNN methods and the current state-of-the-art in Interpretable Deep Reinforcement Learning within the ATARI environment. We believe that our framework serves as a reliable Interpretable Feature Extractor, deepening our understanding of the underlying mechanisms in vision-based Deep Reinforcement Learning models.

\small
\bibliography{bib}
\bibliographystyle{plain}

\end{document}